\documentclass[a4paper,fleqn]{cas-dc}

\usepackage[authoryear]{natbib}

\usepackage{times}
\usepackage{soul}
\usepackage{url}
\usepackage[utf8]{inputenc}
\usepackage[small]{caption}
\usepackage{graphicx}
\usepackage{amsmath}
\usepackage{amsthm}
\usepackage{booktabs}
\usepackage{algorithm}
\usepackage{algorithmic}
\usepackage[switch]{lineno}
\usepackage{tabularx}
\hypersetup{hidelinks}

\usepackage{microtype}
\usepackage{etoolbox}
\usepackage{booktabs}
\usepackage{subcaption}
\usepackage{multirow}
\usepackage{tcolorbox}
\usepackage{refcount}
\usepackage{enumitem}
\usepackage{environ}
\usepackage{pifont}
\usepackage{xspace}
\usepackage{tikz}
\usepackage{fancybox}
\usepackage{amsfonts}
\setitemize{leftmargin=15pt}
\usepackage[export]{adjustbox}

\usepackage[english]{babel}
\usepackage{blindtext}
\usepackage{capt-of}
\usepackage{placeins}

\newcommand{\pname}[1]{\textcolor{black}{GeoReg}}

\newcommand{\todo}[1]{\textcolor{black}{#1}}

\newcommand{\key}[1]{\textcolor{blue}{#1}}
\newcommand{\meta}[1]{\textcolor{orange}{#1}}

\def\tsc#1{\csdef{#1}{\textsc{\lowercase{#1}}\xspace}}
\tsc{WGM}
\tsc{QE}

\begin{document}
\let\WriteBookmarks\relax
\def\floatpagepagefraction{1}
\def\textpagefraction{.001}

\shorttitle{}    

\shortauthors{}  

\title [mode = title]{GeoReg: Weight-Constrained Few-Shot Regression for Socio-Economic Estimation using LLM}  
\tnotetext[2]{This manuscript has been posted as a preprint on arXiv (\href{https://arxiv.org/abs/2507.13323}{arXiv:2507.13323}).}



%

\author[1,6]{Kyeongjin Ahn\fnmark[1]}[type=editor,
      orcid=0009-0001-6109-4093,
]
\ead{kyeongjin.ahn@kaist.ac.kr}
\credit{Conceptualization, Methodology, Validation, Writing}
\author[1]{Sungwon Han} 
\credit{Conceptualization, Writing}
\author[5]{Seungeon Lee} 
\credit{Validation, Writing}
\author[6]{Donghyun Ahn} 
\credit{Data curation}
\author[2]{Hyoshin Kim} 
\credit{Validation}
\author[2]{Jungwon Kim} 
\credit{Validation}
\author[2,1,3]{Jihee Kim} 
\credit{Supervision}
\author[4]{Sangyoon Park} 
\credit{Supervision}
\author[6,1]{Meeyoung Cha\cormark[1]} 
\ead{mia.cha@mpi-sp.org}
\credit{Supervision, Writing, Reviewing}

\cormark[1]

\fnmark[1]



\affiliation[1]{organization={School of Computing, Korea Advanced Institute of Science and Technology (KAIST)},
            city={Daejeon},
            postcode={34141}, 
            country={Republic of Korea}}

\affiliation[2]{organization={College of Business, Korea Advanced Institute of Science and Technology (KAIST)},
            city={Daejeon},
            postcode={34141}, 
            country={Republic of Korea}}

\affiliation[3]{organization={Graduate School of Data Science, Korea Advanced Institute of Science and Technology (KAIST)},
            city={Daejeon},
            postcode={34141}, 
            country={Republic of Korea}}

\affiliation[4]{organization={Division of Social Science, Hong Kong University of Science and Technology (HKUST)},
            city={Kowloon},
            postcode={999077}, 
            country={Hong Kong}}

\affiliation[5]{organization={Max Planck Institute for Software Systems (MPI-SWS)},
            city={Kaiserslautern},
            postcode={67663}, 
            country={Germany}}

\affiliation[6]{organization={Max Planck Institute for Security and Privacy (MPI-SP)},
            city={Bochum},
            postcode={44799}, 
            country={Germany}}

\begin{abstract}
Socio-economic indicators like regional GDP, population, and education levels, are crucial to shaping policy decisions and fostering sustainable development. 
This research introduces \pname{}, a regression model that integrates diverse data sources, including satellite imagery and web-based geospatial information, to estimate these indicators even for data-scarce regions such as developing countries.
Our approach leverages the prior knowledge of large language model to address the scarcity of labeled data, with the language model functioning as a data engineer by extracting informative features to enable effective estimation in few-shot settings.
Specifically, our model obtains contextual relationships between data features and the target indicator, categorizing their correlations as positive, negative, mixed, or irrelevant.
These features are then fed into the linear estimator with tailored weight constraints for each category.
To capture nonlinear patterns, the model also identifies meaningful feature interactions and integrates them, along with nonlinear transformations.
Experiments across three countries at different stages of development demonstrate that our model outperforms baselines in estimating socio-economic indicators, even for low-income countries with limited data availability (Code is provided: \url{https://github.com/kyeongjin0110/GeoReg}). \looseness=-1
\end{abstract}



\maketitle

\section{Introduction}
\label{sec: intro}



Socio-economic indicators, such as economic indicators (e.g., GDP, unemployment rates), demographic statistics (e.g., population figures, birth and death rates), and social indicators (e.g., education levels, access to healthcare), offer crucial data for governments and organizations.
These indicators guide the creation of effective policies. Continuous monitoring of these indicators supports tracking sustainable development progress, identifying inequities, and uncovering vulnerabilities.\looseness=-1

However, constructing such indicators requires substantial financial and human resources, as well as significant time for field surveys and the establishment of administrative systems for data digitization and management. This challenge is particularly pronounced in developing and underdeveloped countries~\citep{benedek2021indicator}. \looseness=-1 
These indicators are not available or reliable at subnational granular levels due to fragmented data collection processes, inconsistent reporting standards, and prioritization of national data over detailed regional statistics~\citep{wenz2023dose}.

\begin{center}
    \includegraphics[width=0.9\columnwidth]{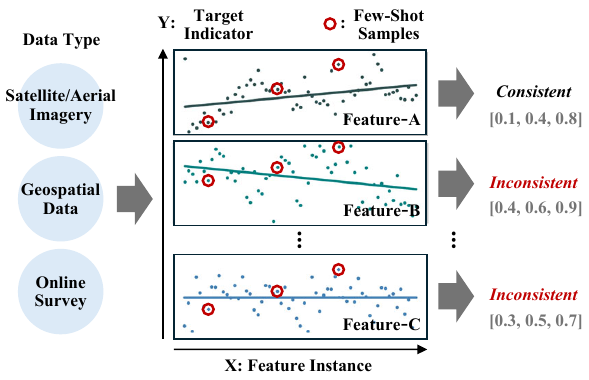}
    \captionof{figure}{Challenges of estimating socio-economic indicators. In few-shot settings, limited samples disrupt finding correct patterns in data. Few-shot samples in Feature-A align with its distribution, while those in Feature-B and Feature-C do not. 
    \looseness=-1}
    \label{fig: intro_image}
\end{center}

Recently, there has been growing interest in using alternative data modalities to predict socio-economic indicators.
Examples include high-resolution satellite and aerial imagery, which are actively explored for their extensive geographical coverage~\citep{albert2017using,ahn2023human}.
In addition to visual data, web-based data, such as geospatial information or official government surveys~\citep{sheehan2019predicting,ren2019predicting}, offers deeper insights into the realities of the field within local contexts.
Combining these diverse data types with AI-based methods enables the estimation of accurate and comprehensive socio-economic indicators. \looseness=-1

However, these emerging AI-driven approaches face several limitations.
One issue is their reliance on the assumption that a large number of ground-truth labels are available for training.
This assumption often does not hold in regions where ground-truth labels are scarce, especially in countries with limited resources for data development, which can hinder both model training and inference.
Figure~\ref{fig: intro_image} shows potential risks with scarce data scenarios, where the available samples may exhibit incorrect distributions that misleads the model to deviate from the actual ground-truth labels.
Another issue is the lack of interpretability in many AI-based methods, which function as black boxes that fail to explain any causal mechanisms behind their predictions. Simply estimating a regional indicator with greater precision may not be enough; revealing the underlying social and economic mechanism is essential to inform and guide effective policy making~\citep{papadakis2024explainable}. \looseness=-1

In this research, we introduce \pname{} that employs a large language model (LLM) as a `data engineer' to extract informative signals from heterogeneous data and socio-economic indicators even under data-scarce conditions.
This approach operates in two key stages: 
Prior to these stages, we define ``modules'' to obtain structured information from various data modalities, such as satellite imagery and geospatial attributes. These modules transform raw input into meaningful features for estimation.
For example, the module ``\texttt{get\_area}'' calculates the area size of a specified region.
In the first stage, leveraging the prior knowledge of LLM, \pname{} determines the most relevant modules to predict a target indicator and uncovers correlations between these modules and the indicator.
In the second stage, we use the selected modules as inputs to train a linear regression model that predicts the target indicator.
The weights of the linear model are constrained to align with the correlations identified by the LLM, ensuring that its knowledge acts as an inductive bias during training to reduce overfitting.
In addition, they are used to discover meaningful feature interactions, which are integrated alongside traditional nonlinear transformations as additional input, enabling the model to effectively capture complex nonlinear patterns.
\looseness=-1

Our model offers several advantages; foremost among them is scalability. The LLM, with its pre-trained knowledge, can extract valuable insights from newly added data sources in addition to original data in predicting broad-ranging socio-economic indicators.
Another merit is interpretability. The linear model allows for a clear explanation of each module's contribution, making it easier to understand the underlying relationships and their implications. This transparency increases confidence in the findings and facilitates communication with researchers and policy makers. \looseness=-1

Experiments in three countries (South Korea, Vietnam, and Cambodia) and multiple indicators (GRDP, Population, and Education indicators) demonstrate that our approach outperforms widely used methods in this field, achieving an average winning rate of 87.2$\%$.
Building on previous efforts in socio-economic indicator estimation, this work makes progress in overcoming data limitations, with the potential to alleviate various social issues, particularly in low-income countries.
\looseness=-1

\section{Related Work}
\label{sec: related}

\subsection{Socio-Economic Indicator Estimation}
Satellite imagery offers broad accessibility for estimating socio-economic indicators. Pioneering work by Jean et al.~\citep{jean2016combining} introduces a CNN-based model to predict poverty, which subsequent research refines for finer tile-level~\citep{han2020learning} and pixel-level~\citep{yeh2020using} predictions. Recent studies propose multi-modal models that integrate external data; for instance, SatelliteBench~\citep{moukheiber2024multimodal} aligns satellite images with public health records, while SATinSL~\citep{suel2021multimodal} supplements vertical satellite perspectives with horizontal ground-level insights from street-view images. \looseness=-1

\subsection{LLMs on Geospatial Data}
Language models increasingly support geospatial inference due to their robust text-processing capabilities. GeoLLM~\citep{manvi2023geollm} estimates indicators like population and asset wealth relying purely on textual information (e.g., addresses), though it lacks visual processing capabilities. To address this, recent models adopt multi-modal architectures. LLaVA~\citep{liu2024visual} applies vision-instruction tuning for visual interactions, while GeoChat~\citep{kuckreja2024geochat} interprets images to answer complex spatial queries, such as object counting and spatial relationship analysis. \looseness=-1

\subsection{Interpretable Socio-Economic Models}
Interpreting socio-economic predictions often relies on post-hoc saliency methods like Grad-CAM~\citep{selvaraju2017grad} to highlight visual features such as buildings and roads~\citep{abitbol2020socioeconomic}. To enhance interpretability, researchers explore external contexts: Sheehan et al.~\citep{sheehan2019predicting} connect Wikipedia data with geographic coordinates, while UrbanClip~\citep{yan2024urbanclip} adopts LLM-generated spatial summaries. However, these approaches primarily provide local post-hoc explanations that offer limited insight into global predictive mechanisms. Additionally, LLM-generated descriptions often require further clarification. \looseness=-1

\begin{figure*}
    \centering
    \includegraphics[height=5.3cm]{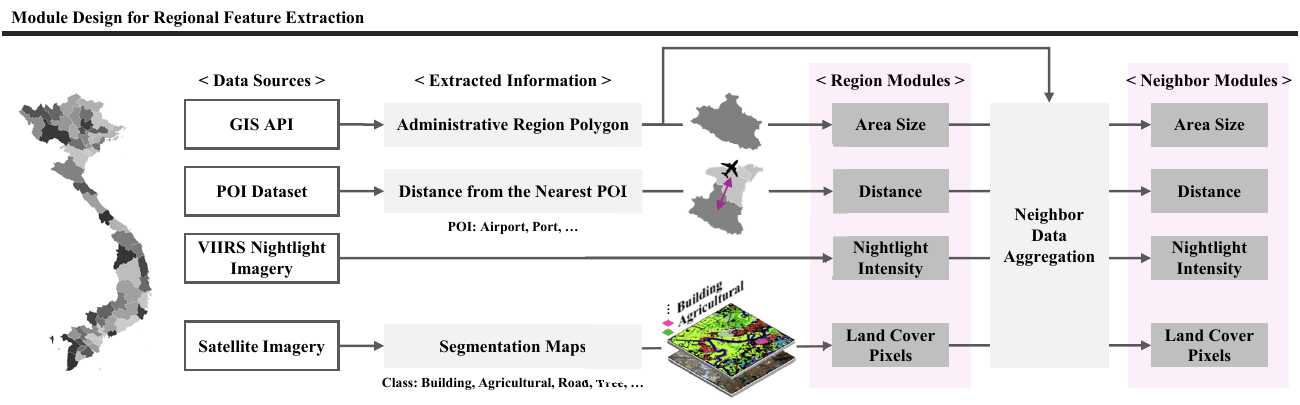}
    \caption{Module design to extract region-aware and neighbor-aware features from heterogeneous data sources for socio-economic indicator estimation. 
    \looseness=-1}
    \label{fig: module_design}    
\end{figure*}

\begin{figure*}
    \centering
    \begin{subfigure}{.559\textwidth}
        \centering
        \raisebox{0mm}{
        \includegraphics[width=1\textwidth]{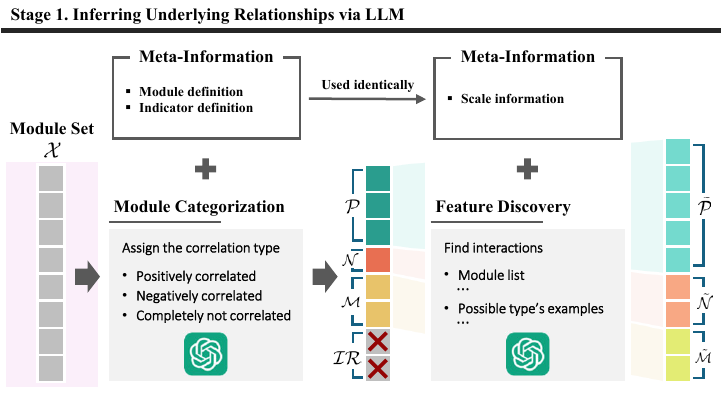}}
        \vspace*{-2mm}
        \label{fig: main_inductive_bias}
    \end{subfigure}
    \hspace{1mm}
    \begin{subfigure}{.412\textwidth}
        \centering
        \raisebox{1mm}{
        \includegraphics[width=1\textwidth]{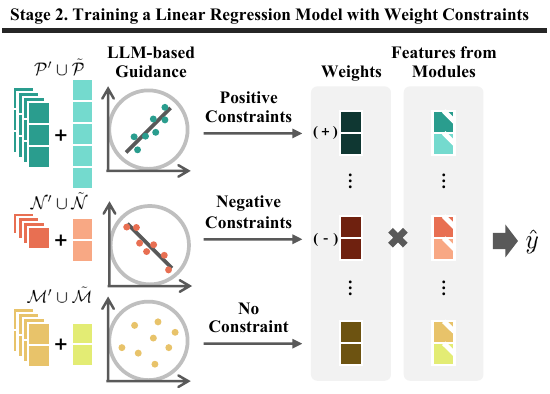}}
        \vspace*{-2mm}
        \label{fig: main_linear_regression}
    \end{subfigure}
    \vspace{-2mm}
    \caption{Overview of \pname{}. 
    In Stage 1, underlying relationships between modules and the target indicator are extracted via LLM by categorizing the module set $\mathcal{X}$ based on relevant meta-information into four groups — \emph{Positive} ($\mathcal{P}$), \emph{Negative} ($\mathcal{N}$), \emph{Mixed} ($\mathcal{M}$), or \emph{Irrelevant} ($\mathcal{IR}$) — and discovering hidden interactions within the categorized subsets. Here, the newly discovered modules in each group are added to their corresponding original ones, which are denoted as $\tilde{\mathcal{P}}$, $\tilde{\mathcal{N}}$, and $\tilde{\mathcal{M}}$, respectively.
    In Stage 2, a linear regression model is trained to estimate the target indicator $\hat{y}$ using the outputs from Stage 1, along with additional augmented sets, including nonlinear transformations (i.e., $\mathcal{P'}$, $\mathcal{N'}$, and $\mathcal{M'}$), guided by distinct weight constraints that reflect their correlations. \looseness=-1}
    \label{fig: model_architecture}
\end{figure*}

\section{Methodology}
\label{sec: method}

\subsection{Problem Statement}
\noindent
\textbf{Problem Definition.} 
Consider $\mathcal{R}$ as a set of regions and $Y$ as the target indicator, where $y_i$ is the ground-truth value of the target indicator for the $i$-th region.
Given that ground-truth values are available for only a few regions during training, the objective of \pname{} is to predict the target indicator value $\hat{y}_i$ such that it closely approximates the corresponding ground-truth value $y_i$. \looseness=-1

\medskip
\noindent
\textbf{Overview.} 
\pname{} is an LLM-based linear regression model for predicting socio-economic indicators in regions with limited training labels.
The process begins with a set of modules designed from heterogeneous data sources to extract features for the given region (i.e., $\mathcal{X}:\mathcal{R}\rightarrow \mathbf{x}$), as illustrated in Figure~\ref{fig: module_design}. 
Figure~\ref{fig: model_architecture} outlines our two-stage approach: LLM-based module categorization (Section~\ref{sec: module_correlation}), followed by a correlation-constrained linear regression (Sections~\ref{sec: weight_constraints}) with nonlinear interactions (Sections~\ref{sec: feature_discovery}).

%
%
\looseness=-1

\subsection{Module Design}  
\label{sec: module_design}
\setlist[itemize]{leftmargin=4mm}
Inspired by socio-economic perspectives~\citep{mellander2015night}, we design modules that extract 26 features from heterogeneous data, such as satellite imagery and geospatial attributes:
\looseness=-1
\begin{itemize}

\item \texttt{get\_area}: Obtains the administrative boundary for a given region and calculates its size. \looseness=-1

\item \texttt{get\_distance\_to\_nearest\_target}: Calculates the distance from a given region's location to the nearest Point of Interest (POI) such as `\textit{airport}' and `\textit{port}', using data from the Natural Earth~\citep{kelso2010introducing}. \looseness=-1

\item \texttt{get\_night\_light}: Extracts cropped VIIRS nightlight images within a given region's boundary and computes the total and average light intensity. \looseness=-1

\item \texttt{count\_area}: Utilizes a pretrained segmentation model~\citep{buscombe2022reproducible} to classify land-cover pixels into eight classes such as `\textit{bareland}', `\textit{rangeland}', `\textit{development}', `\textit{road}', `\textit{tree}', `\textit{water}', `\textit{building}', `\textit{agricultural}', and `\textit{no data}', using data from the OpenEarthMap~\citep{xia2023openearthmap} within a given region's boundary. \looseness=-1

\item \texttt{get\_aggregate\_neighbor\_info}: Identifies neighboring regions that share a boundary point with a given region and aggregates their information based on the outputs of the previously defined modules. \looseness=-1

\end{itemize}
\smallskip
\noindent
The predefined module set denoted as $\mathcal{X}$, where the $j$-th module is represented as $X^{(j)}$.
For a given region $r_i \in \mathcal{R}$, each feature $x^{(j)}_i$ is taken from its corresponding module $X^{(j)}$ (i.e., $x^{(j)}_i = X^{(j)}(r_i)$). \looseness=-1 
\smallskip

\subsection{Knowledge-based Module Categorization}  
\label{sec: module_correlation}
Pre-trained knowledge of language models can help ignore irrelevant or misleading signals and focus on learning nontrivial patterns, particularly in scenarios of a few shots.
This process is supported by module categorization using the predefined module set.
\looseness=-1 \smallskip

\begin{figure}{
\footnotesize
\begin{tcolorbox}[
  colback=black!0!white, colframe=black!20!white, colbacktitle=black!10!white, coltitle=blue!20!black, top=2pt, bottom=2pt ] 
\texttt{Assign the correlation type between \key{<Module>} and \key{<Indicator>} in \key{<Country>}. Here, \meta{<Module Definition>} and \meta{<Indicator Definition>}. Think step by step, and determine one of the following types: \looseness=-1}  \\ 
 \\
\texttt{Type A - Positively correlated} \\
\texttt{Type B - Negatively correlated} \\
\texttt{Type C - Completely not correlated} \\
 \\
\texttt{--- Response ---} \\
\texttt{Explanation:}  \\
\texttt{Answer:}
\end{tcolorbox}
\vspace{-2mm}
\captionof{figure}{Template prompt for module categorization in \pname{}. Key elements are highlighted in blue, with their corresponding meta-information in orange. 
\looseness=-1}
\label{fig: prompt_for_module_categorization}
}
\vspace{-3mm}
\end{figure}

We use LLM to uncover the relationship between each module and the target indicator without relying on a large number of ground-truth labels.
Each module is categorized based on its correlation, $Corr(X^{(j)}, Y)$, between the module ($X^{(j)}$) and the socio-economic indicator ($Y$) using the prompt in Figure~\ref{fig: prompt_for_module_categorization}.
This prompt includes a detailed description of target module and indicator as meta-information, along with explanations of each correlation type.
Our categorization process also adopts the Chain of Thought (CoT) strategy~\citep{wei2022chain} to enable step-by-step reasoning, effectively addressing the complexity of socio-economic estimation tasks. \looseness=-1

Our approach considers four type of correlation categories: \emph{Positive}, \emph{Negative}, \emph{Mixed}, and \emph{Irrelevant}.
A \emph{Positive} correlation indicates that higher values of $X^{(j)}$ correspond to higher values of $Y$, whereas a \emph{Negative} correlation indicates an inverse relationship. A \emph{Mixed} correlation varies across instances, while a \emph{Irrelevant} correlation shows no significant association.
The categorization is repeated five times for reliability, referring to the existing work on LLM self-consistency~\citep{wang2022self}.
The final category for each module is determined by majority votes; if $Corr(X^{(j)}, Y)>0$ appears three or more times, the module is categorized as \emph{Positive} ($\mathcal{P}$); if $Corr(X^{(j)}, Y)<0$ appears three or more times, the module is categorized as \emph{Negative} ($\mathcal{N}$).
In the case of a tie - where $Corr(X^{(j)}, Y)>0$ and $Corr(X^{(j)}, Y)<0$ both appear twice and $Corr(X^{(j)}, Y)=0$ appears once; the module is classified as \emph{Mixed} ($\mathcal{M}$). 
All cases beyond \emph{Positive}, \emph{Negative}, and \emph{Mixed} are considered as \emph{Irrelevant} ($\mathcal{IR}$).

By focusing on categorizing modules based on their general characteristics rather than individual sample values, this process ensures relatively reliable results even in data-scarce scenarios.
Consequently, the data set is formed as $\mathcal{D} = \{ (\mathbf{x}_i, y_i) \}_{i=1}^{N}$, where $\mathbf{x}_i$ contains $N_f$ features of the $i$-th region $r_i$ of selected modules (i.e., $\mathbf{x}_i = \{x^{(j)}_i\}_{j=1}^{N_f}$) and $N$ represents the size of the labeled data. Here, $N << |\mathcal{R}|$. \looseness=-1

\subsection{Linear Regression with Weight Constraints}
\label{sec: weight_constraints}
Linear regression model is computationally efficient and interpretable, which makes it advantageous for socio-economic estimation.
Even with a linear model, a limited number of labels increases the risk of overfitting. To mitigate this, we enforce weight constraints informed by per-module categorization results based on their correlation with the target indicator. 
This approach incorporates the LLM's prior knowledge as an inductive bias, helping prevent overfitting. 
Given a feature vector $\mathbf{x}_i$ and its corresponding ground-truth target indicator value $y_i$ of the $i$-th region $r_i$, the basic linear model is represented using the weight vector $\mathbf{b}$: 
\begin{align}
\begin{split}
    \hat{y}_i & = \mathbf{b} \cdot \mathbf{x}_i + k = \sum_{j=1}^{N_f} \beta^{(j)} x^{(j)}_i + k,
\end{split}
\label{eq: basic_linear_regression}
\end{align}
where $\beta^{(j)}$ is the weight for the $j$-th feature, with $k$ as a bias term. The model parameters are optimized to minimize the mean squared error (MSE) between the predicted value $\hat{y}_i$ and the ground-truth value $y_i$.
Here, \pname{} applies weight constraints based on the correlation of each feature with the target indicator. 
Specifically, features with positive correlations are assigned positive weight constraints, while those with negative correlations are assigned negative constraints.
For features with mixed correlations, no constraints are assigned.
These constraints are defined as follows: \looseness=-1
\begin{align}
  \beta^{(j)} \in 
  \begin{cases}
    \mathbb{R}^{+}, & X^{(j)} \in \mathcal{P} \\
    \mathbb{R}^{-}, & X^{(j)} \in \mathcal{N} \\
    \mathbb{R}, & X^{(j)} \in \mathcal{M}
  \end{cases}
\label{eq: weight_constraints}
\end{align}
This regularization contributes to align the trained weights with LLM's decisions, effectively embedding domain insights into the model.
\looseness=-1

\begin{figure}{
\footnotesize
\begin{tcolorbox}[
  colback=black!0!white, colframe=black!20!white, colbacktitle=black!10!white, coltitle=blue!20!black, top=2pt, bottom=2pt ]
\texttt{Find several new columns related to interactions within the module list for solving the following task. Think step by step for answers. \\ 
\\ 
Task description: Estimate \key{<Indicator>}  in \meta{<Country>} \\ 
\\
Module list: \\
• \key{<Module 1>}: \meta{<Description>} with \meta{<min-max value>} \\
• ... \\
\\ 
Possible types of interaction: \\
• (Module 1)*(Module 2) \\
• ... \\
}
 \\
\texttt{--- Answers ---} \\
\texttt{New column 1: COLUMN $|$ EXPLANATION}  \\
\texttt{New column 2: ...}  
\end{tcolorbox}
\vspace{-2mm}
\captionof{figure}{Template prompt for feature discovery in GeoReg. 
\looseness=-1}
\label{fig: prompt_for_feature_discovery}
}
\vspace{-3mm}
\end{figure}

\subsection{Nonlinear Feature Discovery}
\label{sec: feature_discovery}
Although a linear model is cost-effective and interpretable, it assumes feature independence and cannot accommodate nonlinear patterns.
Therefore, we also consider feature interactions and nonlinear transformations to train the model that capture the hidden relationships in the data.
First, feature interactions are discovered using a prompt in Figure~\ref{fig: prompt_for_feature_discovery}.
To introduce the weight constraints described in Section~\ref{sec: weight_constraints} in the same manner, interactions are generated within each categorized subset.
Interactions from the positive module set $\mathcal{P}$ also positively correlate with the target indicator, while those from the negative module set $\mathcal{N}$ maintain their negative correlations.
These discovered interactions are added to their respective sets, resulting in the extended ones as $\tilde{\mathcal{P}}$, $\tilde{\mathcal{N}}$, and $\tilde{\mathcal{M}}$, respectively.
We utilize the top $k\%$ interactions based on their average Pearson correlation with the original features to filter out outliers.
\looseness=-1

Second, nonlinear transformations, such as
logarithms, square roots, and exponentials, are applied to $\mathcal{P}$, $\mathcal{N}$, and $\mathcal{M}$, to further enrich the feature space.
These transformed features are then combined with their original counterparts, producing augmented sets as $\mathcal{P'}$, $\mathcal{N'}$, and $\mathcal{M'}$, respectively.
The weight constraints of Eq. (\ref{eq: weight_constraints}) are reformulated as follows: \looseness=-1
\begin{align}
  \beta^{(j)} \in 
  \begin{cases}
    \mathbb{R}^{+}, & x^{(j)} \in \mathcal{P'} \cup \tilde{\mathcal{P}} \\
    \mathbb{R}^{-}, & x^{(j)} \in \mathcal{N'} \cup \tilde{\mathcal{N}} \\
    \mathbb{R}, & x^{(j)} \in \mathcal{M'} \cup \tilde{\mathcal{M}}
  \end{cases}
\label{eq: interaction_terms}
\end{align}
The interactions and transformed variations enhance the model's ability to identify potential dependencies, facilitating the representation of intricate economic dynamics. 
After including nonlinear features, we train five models and perform an ensemble by averaging. 
\looseness=-1

\section{Experiments}
\label{sec: experiment}

\begin{figure*}
    \centering
    \begin{subfigure}{.48\textwidth}
        \centering
        \raisebox{0mm}{
        \includegraphics[width=1\textwidth]{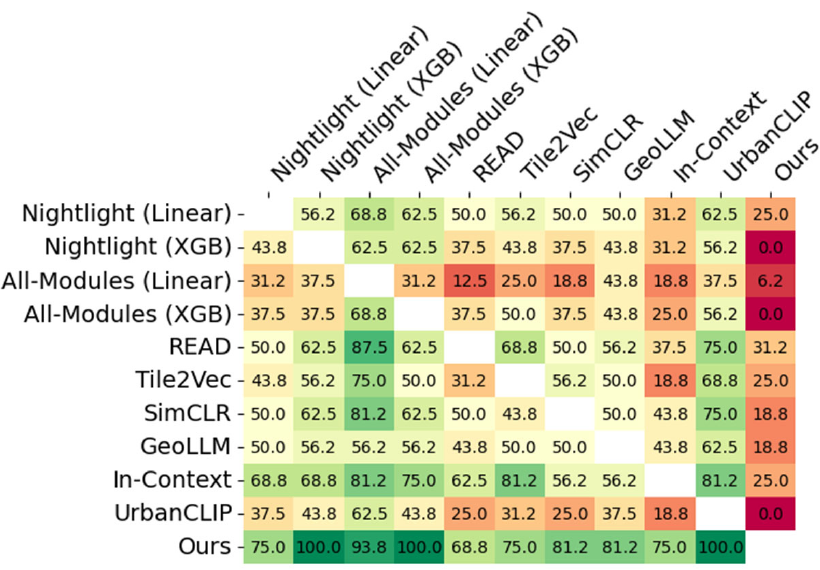}}
        \vspace*{-4mm}
        \caption{Pearson correlation}
        \label{fig: pearson}
    \end{subfigure}
    \begin{subfigure}{.48\textwidth}
        \centering
        \raisebox{0mm}{
        \includegraphics[width=1\textwidth]{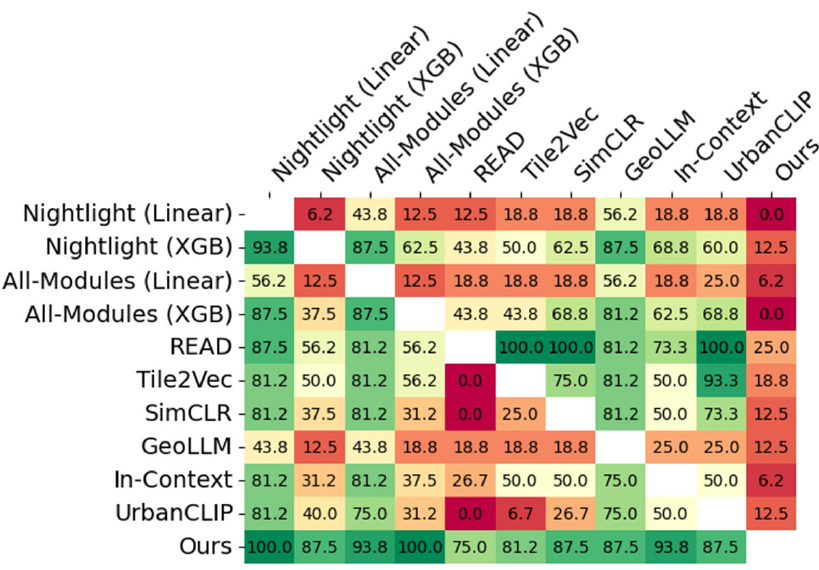}}
        \vspace*{-4mm}
        \caption{RMSE}
        \label{fig: rmse}
    \end{subfigure}
    \caption{Win-matrix summarizing results across different data settings (3 and 5-shot), target indicators (GRDP, POP, and HER), and countries (KOR, VNM, and KHM).
    Darker shades of green represent higher winning rates, while darker shades of red represent lower winning rates. \looseness=-1
    }
    \label{fig: main_results}
\vspace{-3mm}
\end{figure*}

\begin{table*}
\centering
\caption{Performance comparison on the Pearson's correlation and RMSE scores. Results are averaged over repeated experiments with 3-shot and 5-shot settings. The results represent the average across three countries, with the best performances highlighted in bold and cases where our model achieves the second-highest underlined. \looseness=-1
}
{\footnotesize 
\setlength{\tabcolsep}{4pt}       
\renewcommand{\arraystretch}{1.2} 
{
\begin{tabular*}{1.0\textwidth}{@{\extracolsep{\fill}}lcccccccc}
\hline
\multirow{2}{*}{Models} & \multicolumn{4}{c}{Pearson} & \multicolumn{4}{c}{RMSE} \\
\cline{2-9}
 & GRDP & POP & HER & Total & GRDP & POP & HER & Total \\
\hline
(Ablation 1) & 0.591 & 0.514 & 0.345 & 0.483 & 0.916 & 0.840 & 0.052 & 0.603 \\
(Ablation 2) & 0.554 & 0.495 & 0.386 & 0.478 & 0.925 & 0.844 & 0.302 & 0.690 \\
(Ablation 3) & 0.666 & 0.603 & 0.389 & 0.552 & 0.875 & 0.804 & 0.051 & 0.577 \\
(Ablation 4) & 0.594 & 0.567 & 0.374 & 0.512 & 0.918 & \textbf{0.676} & 0.052 & 0.548 \\
(Ablation 5) & 0.662 & 0.550 & 0.340 & 0.518 & 0.879 & 0.833 & 0.051 & 0.588 \\
(Ablation 6) & 0.459 & 0.341 & 0.238 & 0.346 & 1.436 & 1.273 & 0.072 & 0.927 \\
\hline
Ours & \textbf{0.706} & \textbf{0.640} & \textbf{0.405} & \textbf{0.584} &
\textbf{0.816} & \underline{0.763} & \textbf{0.050} & \textbf{0.543} \\
\hline
\end{tabular*}
}
}
\label{table: main_ablation}
\vspace{-3mm}
\end{table*}

\subsection{Experimental Setup}
\noindent
\textbf{Data.} We evaluate three indicators—GRDP (economic), population (demographic), and highly educated ratio (social)—across countries at varied development stages: South Korea (KOR, developed; 229 districts), Vietnam (VNM, growth-stage; 65 provinces), and Cambodia (KHM, developing; 25 provinces).
\looseness=-1
\smallskip

\noindent
\textbf{Implementation details.} Our experiments use GPT-3.5-turbo (temperature 0.5, top-p 1.0) and L2-regularized linear regression. Module categorization is restricted to a maximum of 25 features. Feature discovery utilizes the top 25\% of interactions based on their average Pearson correlation with the original features, supplemented by logarithmic, square-root, and exponential transformations.
\looseness=-1
\smallskip

\noindent
\textbf{Evaluation.} To simulate data-scarce scenarios, we measure Pearson correlation and RMSE under 3 and 5-shot settings, averaging the results across three random runs.
\smallskip
\smallskip

\subsection{Performance Comparison}
\noindent We compare our model against eight baselines grouped into four categories:

\begin{itemize}
\item \textbf{Traditional Regression (Linear \& XGBoost):} \textbf{Nightlight}~\citep{bagan2015analysis} estimates targets using regional luminosity (average and sum), while \textbf{All-Modules} utilizes the entire unselected feature set.

\item \textbf{Visual Representation Models:} These models extract embeddings from satellite images to train a subsequent linear regressor. We evaluate \textbf{READ}~\citep{han2020lightweight} (weakly supervised learning), \textbf{Tile2Vec}~\citep{jean2019tile2vec} (unsupervised learning), and \textbf{SimCLR}~\citep{chen2020simple} (contrastive learning).

\item \textbf{LLM-based Models (GPT-3.5-turbo):} \textbf{GeoLLM}~\citep{manvi2023geollm} is fine-tuned using regional addresses and nearby locations. Alternatively, \textbf{In-Context Learning}~\citep{brown2020language} operates via few-shot text paragraphs generated from all module set.

\item \textbf{Vision-Language Models (VLMs):} \textbf{UrbanCLIP}~\citep{yan2024urbanclip} utilizes multi-modal embeddings that align satellite imagery with regional text descriptions.
\end{itemize}
\looseness=-1
\smallskip

\noindent
\textbf{Comparison Results.} 
To validate the model's consistent effectiveness under various scenarios, a win-matrix is used to map the winning rates of x-axis models against y-axis baselines across data settings (3 and 5-shot), indicators (GRDP, POP, and HER), and countries (KOR, VNM, and KHM). Complete results are in \todo{\ref{sec: appendix_full_results}}.
\looseness=-1

Figure~\ref{fig: main_results} presents our model's robust performance, achieving an 87.2\% average winning rate over all baselines across both Pearson correlation and RMSE metrics. Compared to \textbf{traditional regressions} (Nightlight, All-Modules), our approach demonstrates the value of integrating heterogeneous data. Notably, while Nightlight yields impressive results using a single feature—justifying its inclusion as a core module—it cannot match our comprehensive framework. Furthermore, our model consistently outperforms \textbf{visual representation models} (READ, Tile2Vec, SimCLR), indicating that vision-only embeddings are insufficient for capturing complex socio-economic patterns. We also observe clear advantages over \textbf{LLM-based models} (GeoLLM, In-Context Learning), confirming that explicitly structuring LLM insights performs better than relying purely on implicitly embedded knowledge. Finally, in comparison with the \textbf{VLM-bsed model} (UrbanCLIP), our dedicated modules prove more effective; while VLMs often abstract away fine-grained details during querying, our approach explicitly preserves vital detailed insights, such as nightlight intensity and land cover ratios. \looseness=-1

\subsection{Ablation Study}
\noindent\textbf{Component Analysis.} Table~\ref{table: main_ablation} reveals the impact of our module categorization (weight constraints) and feature discovery process. We explore the following variations:
\textsf{(Ablation 1: simple linear)} baseline using the features from entire module set;
\textsf{(Ablation 2: feature selection)} LLM-filtered features only;
\textsf{(Ablation 3: weight constraints only)} categorization-based constraints without nonlinear features;
\textsf{(Ablation 4: feature discovery only)} LLM-discovered interactions without weight constraints;
\textsf{(Ablation 5: polynomial)} features from entire module set with all second-degree polynomial terms~\citep{ostertagova2012modelling}; and
\textsf{(Ablation 6: AutoFeat)} features from entire module set with AutoFeat~\citep{horn2020autofeat}. \looseness=-1

Results indicate that simply removing irrelevant modules (Ablation 2) fails to improve performance due to data scarcity. In contrast, applying weight constraints (Ablation 3) successfully guides the learning process by emphasizing relevant information. Furthermore, introducing LLM-generated feature interactions (Ablation 4) captures complex patterns better than baselines (Ablations 1--2). While exhaustive polynomial expansion (Ablation 5) and the AutoFeat (Ablation 6) yield comparable accuracy by generating hundreds of combinations, our LLM-driven discovery achieves similar results with significantly lower computational complexity. Ultimately, both components are essential for maximizing performance while minimizing overhead. \looseness=-1
\smallskip

\noindent
\textbf{Hyperparameter Analysis.}
We study how the number of feature interactions ($k\%$) and the ensemble size affect the performance of the model. 
Using a feature interaction of $25\%$ and an ensemble size of $5$ achieve an optimal balance between efficiency and performance. 
\looseness=-1

\begin{figure*}
    \centering
    \begin{subfigure}{.3\textwidth}
        \centering
        \includegraphics[width=1\textwidth]{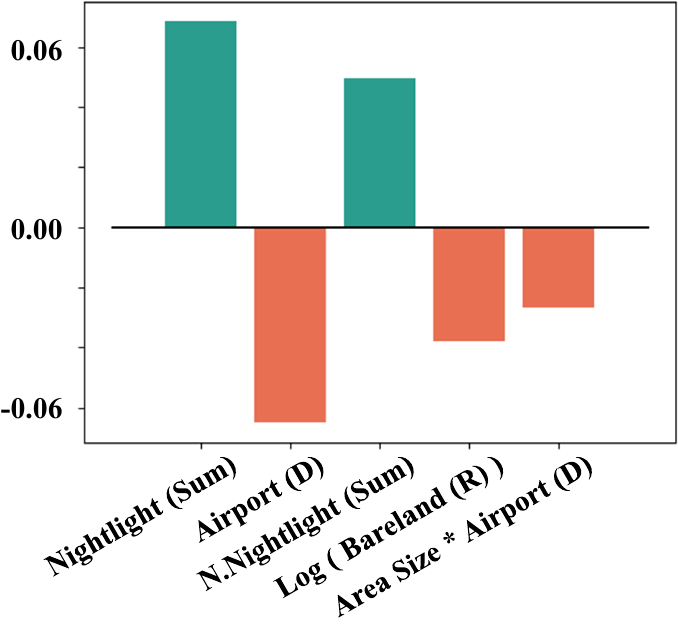}
        \caption{KOR}
        \label{fig: interpretability_KOR}
    \end{subfigure}
    \hspace{2mm}
    \begin{subfigure}{.3\textwidth}
        \centering
        \includegraphics[width=1\textwidth]{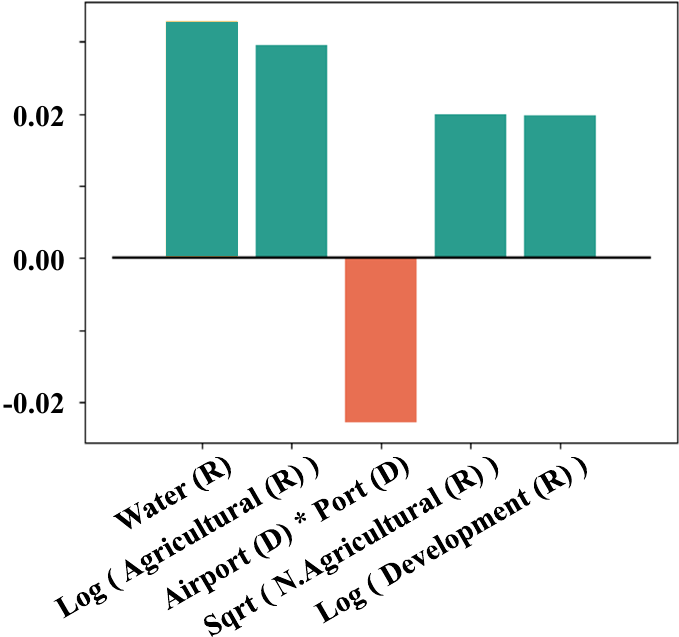}
        \caption{KHM}
        \label{fig: interpretability_KHM}
    \end{subfigure}
    \caption{Top five learned weights from \pname{}, trained to predict the POP indicator of KOR and KHM. The module names are shown on the x-axis, while the values of the learned weights on the y-axis. Bar colors indicate module categories: green for positive, red for negative, and yellow for mixed module sets. \looseness=-1}
    \label{fig: discussion_interpretability}
\end{figure*}

\section{Discussion}
\label{sec: discussion}

\noindent
\textbf{\emph{Q1. Do the learned weights meaningfully explain the predicted target indicator?}}
We present a case study that demonstrates the insights from the learned weights in \pname{}. Figure~\ref{fig: discussion_interpretability} shows the top five weights of our model, ranked by absolute magnitude, trained to predict the POP indicator for South Korea (KOR) and Cambodia (KHM). These results highlight notable differences in the learned weights between a developed country and a developing country. \looseness=-1

In South Korea, regions tend to be densely populated when night time lights are brighter and less populated when they are farther from an airport. In contrast, in Cambodia, agriculture-related variables play a key role in estimating population, reflecting the industrial structure of developing countries. Interestingly, the water-related variable is important for population estimation in Cambodia. We hypothesize that this may reflect the lasting influence of Angkor Wat and its historical water infrastructure, which once sustained dense settlements and continues to shape regional development through tourism~\citep{kummu2009water}. 
Our result shows how the interpretability of our model can provide valuable insights by revealing the relative importance of different features in the prediction of the population. Although our result does not imply a causal relationship, it can offer useful perspectives for policy makers. We expect this interpretability to be valuable at the local level, especially for regions with limited data conditions, as diverse and unique underlying economic mechanisms can be found in local economies and communities. 
\looseness=-1
\smallskip

\begin{figure}
    \centering
    \begin{subfigure}{.235\textwidth}
        \centering
        \raisebox{0mm}{
        \includegraphics[width=1\textwidth]{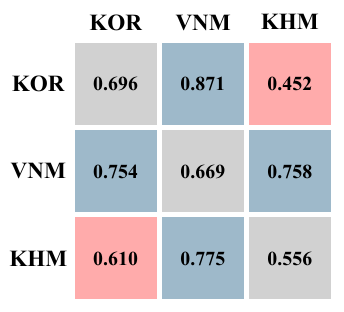}}
        \caption{POP}
        \label{fig: transferability_POP}
    \end{subfigure}
    \begin{subfigure}{.235\textwidth}
        \centering
        \raisebox{0mm}{
        \includegraphics[width=1\textwidth]{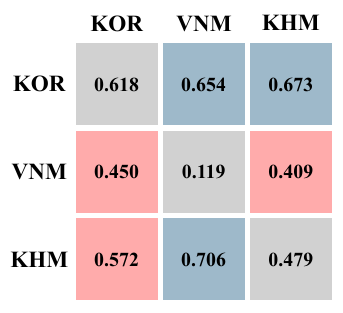}}
        \caption{HER}
        \label{fig: transferability_HER}
    \end{subfigure}
    \caption{Cross-country transferability. Pearson correlation matrices are shown for (a) POP and (b) HER, averaged over 3 and 5-shot within-country (diagonal) and full-shot across-country (off-diagonal) comparisons. Blue indicates higher transferability than within-country results, while red indicates lower. \looseness=-1}
    \label{fig: discussion_transferability}
\end{figure}

\noindent
\textbf{\emph{Q2. Can the model be transferred to different countries?}}
To examine the transferability of the model, we analyze the Pearson correlation for the POP and HER indicators in the designated source-target country pairs.
Here, the source country refers to the one used for training, while the target country refers to the one used for evaluation. 
Figure~\ref{fig: discussion_transferability} shows the results, with each matrix displaying the source countries on the x-axis and the target countries on the y-axis.
The POP indicator exhibits higher transferability than the HER indicator, which may be because the data distribution of the HER indicator varies more between countries at different stages of development compared to that of the POP indicator.
VNM consistently achieves high Pearson correlation values as a source country for both indicators, potentially reflecting its unique position as a bridge between developed and developing countries. \looseness=-1
\smallskip

\noindent
\textbf{\emph{Q3. Are the results of the LLM reliable?}}
To quantitatively assess the reliability of the LLM's results, we perform Jaccard similarity analysis for module categorization and mutual information (MI) analysis for feature discovery. \looseness=-1
\smallskip

\noindent\textbf{Reliability of Categorization Task.}  
To construct this ground-truth, we compute the Pearson correlation between each feature and the target indicator, then classify modules into one of three correlation types — \emph{Positive} ($\mathcal{P}$), \emph{Negative} ($\mathcal{N}$), and \emph{Mixed} ($\mathcal{M}$) — based on a threshold $\tau$.
A module is labeled \emph{Positive} if its Pearson correlation value exceeds $\tau$, \emph{Negative} if below $-\tau$, and  \emph{Mixed} if within $[-\tau, \tau]$.
Table~\ref{table: similarity_analysis} presents the Jaccard similarity scores for the POP indicator across countries. 
For each type of correlation within a country, the scores are averaged over different threshold values, $\tau \in \left\{0.05, 0.10, 0.15 \right\}$.
In particular, cases such as (a) KOR and (c) KHM achieve reliable scores in both the \emph{Positive} and \emph{Negative} module sets, underscoring the robustness of the LLM-guided categorization approach. \looseness=-1
\smallskip

\begin{figure}
    \centering
    \hspace{-6mm}
    \begin{subfigure}{.25\textwidth}
        \centering
        \raisebox{0mm}{
        \includegraphics[width=1\textwidth]{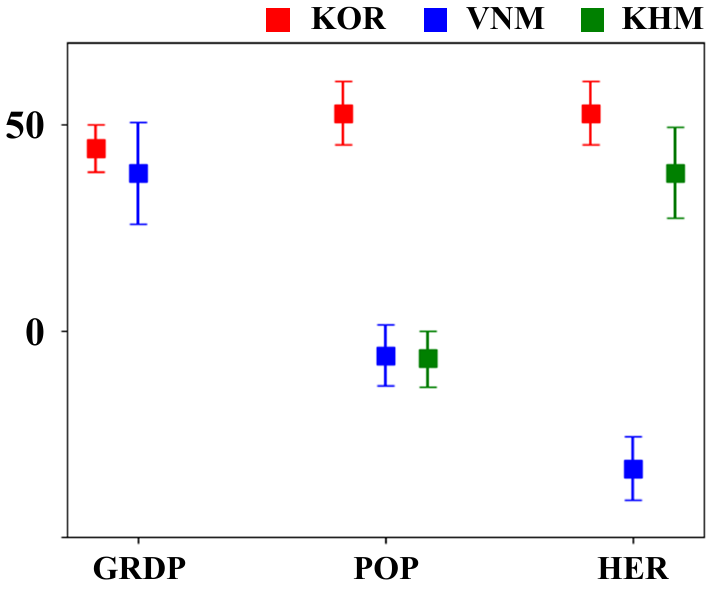}}
        \vspace*{-4mm}
        \caption{3-Shot}
        \label{fig: interaction_usefulness_fewshot_3}
    \end{subfigure}
    \begin{subfigure}{.25\textwidth}
        \centering
        \raisebox{0mm}{
        \includegraphics[width=1\textwidth]{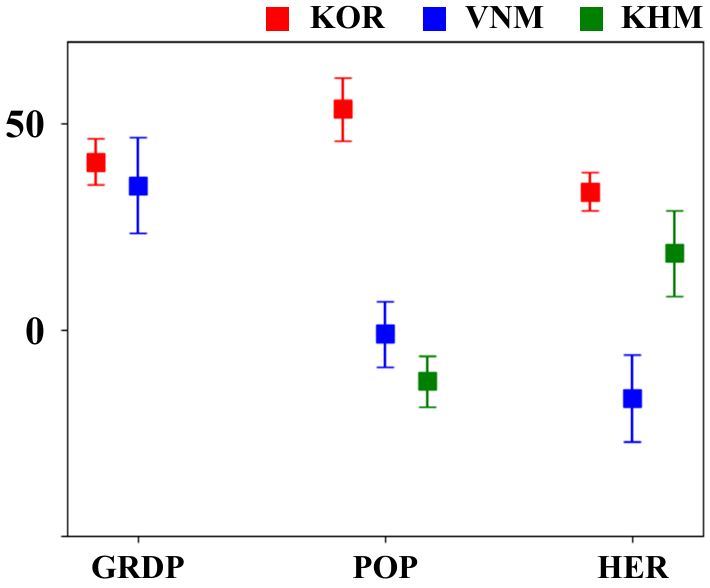}}
        \vspace*{-4mm}
        \caption{5-Shot}
        \label{fig: interaction_usefulness_fewshot_5}
    \end{subfigure}
    \caption{Analysis of LLM-based feature discovery reliability through mutual information (MI) measurement. The MI difference mean between the discovered and original features, along with its standard error, is presented for each indicator across three countries, shown for 3-shot (a) and 5-shot (b) settings, respectively. \looseness=-1}
    \label{fig: interaction_usefulness}
\vspace{-3mm}
\end{figure}

\noindent\textbf{Reliability of Discovery Task.} We evaluate the effectiveness of feature discovery using mutual information (MI), which quantifies the relationship between features and the ground-truth target indicator.
For comparison, the percentage difference between each discovered interaction feature's MI and the average MI of the original features is computed.
These differences are then averaged across all interaction features to derive a metric that we refer to as \emph{MI difference mean}.
The MI difference mean, along with its standard error, is reported for each indicator in the countries in Figure~\ref{fig: interaction_usefulness}. \looseness=-1

\begin{table}[t]
\centering
\caption{Analysis of LLM-based feature discovery reliability through mutual information (MI) measurement. The MI difference mean between the discovered and original features, along with its standard error, is presented for each indicator across three countries, shown for 3-shot (a) and 5-shot (b) settings, respectively.}
\label{table: similarity_analysis}

{\footnotesize
\setlength{\tabcolsep}{3.8pt}      
\renewcommand{\arraystretch}{1.2}

\begin{tabular}{@{}ccc ccc ccc@{}}
\hline
\multicolumn{3}{c}{KOR} & \multicolumn{3}{c}{VNM} & \multicolumn{3}{c}{KHM} \\
\hline
$\mathcal{P}$ & $\mathcal{N}$ & $\mathcal{M}$ &
$\mathcal{P}$ & $\mathcal{N}$ & $\mathcal{M}$ &
$\mathcal{P}$ & $\mathcal{N}$ & $\mathcal{M}$ \\
\hline
0.819 & 0.602 & 0.697 & 0.457 & 0.572 & 0.600 & 0.523 & 0.630 & 0.286 \\
\hline
\end{tabular}
}
\end{table}

In most cases, the MI difference mean is significantly positive, indicating that the discovered features capture more information than the original ones.
While some cases exhibit a negative mean MI difference, this does not necessarily imply that the discovered features are devoid of useful information.
Instead, they may capture unique information that is not present in the original ones, even if the overall quantity of MI is smaller.
To validate this, we compare the model's performance with and without feature discovery by analyzing the Pearson correlation.
Even in cases where the MI difference mean is negative, the model remains robust, often improving when feature interactions are applied.
For example, in Vietnam (VNM) and Cambodia (KHM) for the POP indicator, incorporating feature interactions led to performance gains of 0.15 and 0.10, respectively, averaged over 3 and 5-shot settings. \looseness=-1

\section{Conclusion}
\label{sec: conclusion}

This paper presents \pname{}, a regression model that uses the prior knowledge from LLM based on satellite imagery and web-based information to estimate key socio-economic indicators in data-scarce scenarios.
By categorizing data features based on their correlations with the target indicator using the LLM, our approach integrates domain-informed priors through weight constraints, guiding the model toward relevant patterns and reducing the risk of overfitting in few-shot settings.
Furthermore, \pname{} explores interactions within features, capturing complex patterns that go beyond the initial straightforward attributes of the data.
Extensive experiments validate the model's effectiveness across a range of indicators and countries, while our discussion delves into its potential for broader applications.
\looseness=-1
\smallskip

\section*{Acknowledgements}
This work was supported by the Max Planck Institute for Security and Privacy, and by the National Research Foundation of Korea (NRF) through grants funded by the Korean government (MSIT) (No. RS-2022-00165347 and RS-2025-00563196).
\looseness=-1
\smallskip

\clearpage 





\appendix
\smallskip




\section{Complete Results}
\label{sec: appendix_full_results}
\noindent We provide the complete results in Table~\ref{table: appendix_pearson} and Table~\ref{table: appendix_rmse}.
\looseness=-1

\begin{table*}
\centering
\vspace{-1.5mm}
\caption{Detailed Pearson correlation results for comparison with baselines.}
\label{table: appendix_pearson}

{\footnotesize
\setlength{\tabcolsep}{6pt}  
\renewcommand{\arraystretch}{1.2} 

\begin{tabularx}{\textwidth}{@{}X cccc cccc ccc@{}}
\hline
\multirow{2}{*}{Models} &
\multicolumn{4}{c}{South Korea} &
\multicolumn{4}{c}{Vietnam} &
\multicolumn{3}{c}{Cambodia} \\
\cline{2-12}
 & GRDP & POP & HER & Total & GRDP & POP & HER & Total & POP & HER & Total \\
\hline
Nightlight (Linear)      & 0.552 & 0.540 & 0.435 & 0.509 & 0.502 & 0.487 & -0.197 & 0.264 & 0.114 & 0.807 & 0.461 \\
Nightlight (XGB)         & 0.421 & 0.407 & 0.265 & 0.364 & 0.733 & 0.550 & 0.045  & 0.442 & 0.168 & 0.121 & 0.144 \\
All-Modules (Llinear)    & 0.333 & 0.353 & 0.361 & 0.349 & 0.518 & 0.499 & -0.199 & 0.273 & 0.079 & 0.274 & 0.177 \\
All-Modules (XGB)        & 0.479 & 0.529 & 0.502 & 0.504 & 0.337 & 0.286 & -0.038 & 0.195 & 0.215 & 0.373 & 0.294 \\
READ                     & 0.459 & 0.509 & 0.599 & 0.522 & 0.386 & 0.318 & 0.220  & 0.308 & \textbf{0.636} & 0.398 & 0.517 \\
Tile2Vec                 & 0.327 & 0.406 & 0.418 & 0.384 & 0.418 & 0.389 & 0.154  & 0.321 & 0.621 & 0.356 & 0.489 \\
SimCLR                   & 0.503 & 0.538 & 0.580 & 0.540 & 0.358 & 0.367 & 0.164  & 0.296 & 0.568 & 0.324 & 0.446 \\
GeoLLM                   & 0.099 & 0.465 & 0.463 & 0.342 & 0.501 & 0.602 & 0.252  & 0.452 & 0.558 & -0.077 & 0.241 \\
In-Context Learning      & 0.551 & 0.352 & 0.467 & 0.457 & 0.631 & 0.498 & \textbf{0.305} & 0.478 & 0.447 & \textbf{0.855} & \textbf{0.651} \\
UrbanCLIP                & 0.398 & 0.354 & 0.234 & 0.329 & 0.445 & 0.401 & -0.039 & 0.269 & 0.543 & 0.271 & 0.407 \\
\hline
Ours & \textbf{0.666} & \textbf{0.696} & \textbf{0.618} & \textbf{0.660} &
\textbf{0.746} & \textbf{0.669} & 0.119 & \textbf{0.511} &
0.556 & 0.479 & \underline{0.517} \\
\hline
\end{tabularx}
}

\vspace{2mm}
\end{table*}

\begin{table*}
\centering
\vspace{-1.5mm}
\caption{Detailed RMSE results for comparison with baselines.}
\label{table: appendix_rmse}

{\footnotesize
\setlength{\tabcolsep}{6pt}  
\renewcommand{\arraystretch}{1.2} 

\begin{tabularx}{\textwidth}{@{}X cccc cccc ccc@{}}
\hline
\multirow{2}{*}{Models} &
\multicolumn{4}{c}{South Korea} &
\multicolumn{4}{c}{Vietnam} &
\multicolumn{3}{c}{Cambodia} \\
\cline{2-12}
 & GRDP & POP & HER & Total & GRDP & POP & HER & Total & POP & HER & Total \\
\hline
Nightlight (Linear)      & 1.546 & 1.728 & 0.239 & 1.171 & 1.725 & 0.716 & 0.038 & 0.826 & 14.357 & 0.274 & 7.316 \\
Nightlight (XGB)         & 1.101 & 1.095 & 0.127 & 0.775 & 0.681 & 0.536 & 0.028 & 0.415 & 1.069  & 0.038 & 0.554 \\
All-Modules (Llinear)    & 2.103 & 2.313 & 0.205 & 1.540 & 0.988 & 0.530 & 0.051 & 0.523 & 2.932  & 0.062 & 1.497 \\
All-Modules (XGB)        & 1.112 & 1.055 & 0.107 & 0.758 & 0.941 & 0.614 & 0.030 & 0.528 & 1.109  & 0.034 & 0.571 \\
READ                     & 1.227 & 1.130 & 0.098 & 0.818 & 0.866 & 0.600 & \textbf{0.027} & 0.498 & 0.929 & 0.036 & 0.482 \\
Tile2Vec                 & 1.342 & 1.251 & 0.112 & 0.901 & 0.875 & 0.603 & \textbf{0.027} & 0.502 & 0.954 & 0.036 & 0.495 \\
SimCLR                   & 1.374 & 1.286 & 0.119 & 0.926 & 0.936 & 0.632 & \textbf{0.027} & 0.532 & 1.004 & 0.037 & 0.521 \\
GeoLLM                   & 6.369 & 2.697 & \textbf{0.083} & 3.050 & 0.956 & 0.543 & 0.031 & 0.510 & 4.796 & 9.163 & 6.979 \\
In-Context Learning      & 1.240 & 1.883 & 0.102 & 1.075 & 0.778 & 0.710 & 0.028 & 0.506 & 1.885 & 0.036 & 0.961 \\
UrbanCLIP                & 1.444 & 1.362 & 0.126 & 0.977 & 0.965 & 0.649 & 0.027 & 0.547 & 0.992 & 0.037 & 0.514 \\
\hline
Ours & \textbf{0.937} & \textbf{0.858} & \underline{0.091} & \textbf{0.629} &
\textbf{0.695} & \textbf{0.516} & \underline{0.028} & \textbf{0.413} &
\textbf{0.914} & \textbf{0.032} & \textbf{0.473} \\
\hline
\end{tabularx}
}

\vspace{2mm}
\end{table*}

\printcredits

\bibliographystyle{cas-model2-names}

\bibliography{cas-refs}




\end{document}